\documentclass[conference]{IEEEtran}
\IEEEoverridecommandlockouts
\usepackage{cite}
\usepackage{amsmath,amssymb,amsfonts}
\usepackage{algorithmic}
\usepackage{graphicx}
\usepackage{textcomp}
\usepackage{xcolor}
\usepackage{tikz}
\usepackage{booktabs}
\usepackage{multirow}

\usepackage{subcaption}
\usetikzlibrary{shapes,arrows,positioning}
\def\BibTeX{{\rm B\kern-.05em{\sc i\kern-.025em b}\kern-.08em
    T\kern-.1667em\lower.7ex\hbox{E}\kern-.125emX}}
    
\begin{document}

\title{Guardrailed Uplift Targeting: A Causal Optimization Playbook for Marketing Strategy}

\author{\IEEEauthorblockN{Deepit Sapru}
\IEEEauthorblockA{University of Illinois Urbana-Champaign\\
Email: dsapru2@illinois.edu}
}
\maketitle

\begin{abstract}
This paper introduces a marketing decision framework that optimizes customer targeting by integrating heterogeneous treatment effect estimation with explicit business guardrails. The objective is to maximize revenue and retention while adhering to constraints such as budget, revenue protection, and customer experience. The framework first estimates Conditional Average Treatment Effects (CATE) using uplift learners, then solves a constrained allocation problem to decide whom to target and which offer to deploy. It supports decisions in retention messaging, event rewards, and spend-threshold assignment. Validated through offline simulations and online A/B tests, the approach consistently outperforms propensity and static baselines, offering a reusable playbook for causal targeting at scale.
\end{abstract}


\section{Introduction}
Modern e-commerce platforms face the critical challenge of optimizing customer targeting strategies while managing business constraints. Traditional approaches often rely on propensity scoring or random assignment, which can lead to suboptimal outcomes and inefficient resource allocation \cite{ascarza2018, devriendt2021}. The fundamental limitation of these methods lies in their inability to account for heterogeneous treatment effects and business constraints simultaneously.

Customer engagement initiatives, including promotions, discounts, and retention campaigns, represent significant investments for e-commerce businesses. However, without proper targeting mechanisms, these investments may fail to achieve desired returns or even negatively impact customer behavior \cite{berger2018, bauer2019}. Previous research has demonstrated that targeting customers based solely on propensity scores can be counterproductive, particularly in retention scenarios where high-risk customers may respond negatively to interventions \cite{ascarza2018}.

The emergence of causal machine learning has enabled more sophisticated approaches to treatment effect estimation. Uplift modeling techniques, including meta-learners and forest-based estimators, provide methods for estimating individual treatment effects \cite{kunzel2019, wager2018}. Recent work has demonstrated the value of combining uplift modeling with optimization frameworks in e-commerce settings, including knapsack formulations for promotion personalization \cite{albert2022} and retrospective uplift modeling within ROI constraints \cite{goldenberg2020}. Extensions to multiple treatments with explicit cost optimization have further advanced practical applicability \cite{zhao2019}. However, translating these estimates into actionable targeting strategies that systematically respect diverse business constraints remains challenging.

This paper addresses this gap by introducing a comprehensive framework that combines uplift modeling with constrained optimization. Our approach enables marketers to identify optimal customer segments for targeting while adhering to practical business constraints such as budget limitations, revenue protection, and customer experience metrics. The framework provides a systematic methodology for converting causal estimates into executable marketing strategies.

We validate our approach through multiple large-scale experiments across different business scenarios, including customer retention, revenue maximization, and promotional offer optimization. The results demonstrate consistent improvements over baseline approaches in both offline evaluations and online A/B testing environments. Our contributions include a generalized targeting framework, practical implementation guidelines, and empirical validation across diverse e-commerce applications.

The remainder of this paper is organized as follows: Section 2 reviews related work in uplift modeling and constrained optimization. Section 3 details our methodological framework. Section 4 describes the simulation setup for offline validation. Section 5 presents experimental applications and results. Section 6 discusses implementation considerations, and Section 7 concludes with implications for marketing practice.

\subsection{Contributions}
Our main contributions are:
\begin{itemize}
    \item A two-stage \emph{guardrailed uplift targeting} framework that combines CATE estimation with constrained optimization for practical marketing decisions.
    \item A detailed simulation setup for offline validation, including data generation, treatment assignment, response functions, and guardrail constraints.
    \item Empirical validation across three business scenarios (retention, event revenue, spend thresholds) showing consistent improvement over baselines.
    \item Implementation guidelines covering data requirements, model selection, constraint specification, and computational scalability.
    \item Publicly available code and a reusable playbook for marketers to operationalize causal targeting with guardrails.
\end{itemize}

\section{Related Work}

Our framework for guardrailed uplift targeting intersects with several streams of research in machine learning, data systems, and optimization. Foundational work in uplift modeling established core methodologies for treatment effect estimation, including comprehensive surveys of the field \cite{gutierrez2017} and the development of specialized decision tree algorithms for single and multiple treatment scenarios \cite{rzepakowski2012}. While traditional uplift modeling has focused primarily on treatment effect estimation \cite{devriendt2021}, our approach extends this by integrating business constraints into the optimization process.

Recent work by \cite{paul2024bias} addresses model fairness through gender stereotype-aware loss regularizers, highlighting the importance of ethical considerations in ML systems—a concern that aligns with our use of business guardrails to prevent negative customer experiences. Similarly, \cite{paul2024sentiment} and \cite{paul2024fakenews} explore model robustness through rule-based feature extraction and embedding comparisons, respectively, underscoring the broader need for reliable and interpretable models in production environments.

In the realm of data infrastructure, \cite{arora2025rdf} proposes RDF-based structures for efficient querying in decentralized systems, while \cite{arora2025entity} introduces explainable graph learning for entity resolution. Both contributions emphasize scalable and transparent data management, which is foundational for any enterprise-level targeting system. Further, \cite{maheshwari2025} introduces efficient dictionary structures with working-set properties, offering computational optimizations relevant to our large-scale constraint solving.

On the architectural side, \cite{mamtani2025} improves vision transformers via feature map regularization, illustrating how model-level innovations can enhance downstream task performance. Complementary to this, \cite{paul2024ocr} compares convolutional and feedforward networks, and \cite{paul2024recommendation} integrates tags and latent factors for better recommendations—each contributing to the model selection and personalization strategies we employ.

Finally, broader system-level challenges are addressed by \cite{thomas2025privacy} in balancing privacy with utility, by \cite{thomas2025software} in maintaining evolvable software architectures, and by \cite{thomas2025healthcare} in integrating fragmented data sources. These reflect the same trade-offs between performance, compliance, and sustainability that our guardrailed framework seeks to navigate.

Together, these works inform the multi-stage, constraint-aware approach we propose, bridging causal ML with practical deployment requirements.

\section{Methodological Framework}

\subsection{Problem Formulation}
We consider the problem of optimal customer targeting under business constraints. Let $C = \{1, 2, \ldots, N\}$ represent the set of customers and $\mathcal{T} = \{t_0, t_1, \ldots, t_K\}$ denote available treatments, where $t_0$ typically represents a control or baseline condition. For each customer $i$, we observe covariates $X_i \in \mathcal{X}$, assigned treatment $T_i \in \mathcal{T}$, and outcome $Y_i \in \mathbb{R}$.

The objective is to learn a targeting policy $\pi: \mathcal{X} \to \mathcal{T}$ that maximizes the expected outcome while satisfying business constraints. Formally, we seek to solve:
\begin{equation}
\max_{\pi} \mathbb{E}[Y(\pi(X))] \quad \text{subject to} \quad g_j(\pi) \leq c_j \quad \forall j \in J
\end{equation}
where $g_j$ represent constraint functions and $c_j$ are corresponding limits.

The policy $\pi$ can be represented as a binary matrix where $\pi_{ik} = 1$ if treatment $t_k$ is assigned to customer $i$, and 0 otherwise, with $\sum_k \pi_{ik} = 1$ ensuring each customer receives exactly one treatment.

\subsection{Uplift Estimation Stage}
The first stage of our framework involves estimating Conditional Average Treatment Effects (CATE) using uplift models. For a binary treatment case with $t_0$ as control and $t_1$ as treatment, we define CATE as:
\begin{equation}
\tau(x_i) = \mathbb{E}[Y_i(1) | X_i = x_i] - \mathbb{E}[Y_i(0) | X_i = x_i]
\end{equation}
where $Y_i(1)$ and $Y_i(0)$ represent potential outcomes under treatment and control, respectively. This notation follows the potential outcomes framework formalized by Rubin \cite{rubin2005}, which provides the theoretical foundation for causal inference from experimental and observational data.

We employ several state-of-the-art uplift estimators in our framework:
\begin{itemize}
\item \textbf{Causal Forest}: An adaptation of random forests for treatment effect estimation that handles high-dimensional covariates effectively \cite{wager2018}. Related approaches include orthogonal random forests, which achieve improved convergence properties through Neyman-orthogonal score functions \cite{oprescu2019}.
\item \textbf{Double Machine Learning}: A doubly robust method that combines outcome and propensity score modeling to reduce bias \cite{chernozhukov2018}.
\item \textbf{Forest Doubly Robust Learner}: Integrates doubly robust estimation with forest-based methods for improved performance \cite{zhang2020}.
\end{itemize}

The unconfoundedness assumption is crucial for identifying causal effects from observational data. This assumption requires that treatment assignment is independent of potential outcomes conditional on observed covariates. While this assumption is fundamentally untestable, careful experimental design and covariate selection can make it more plausible.

\begin{figure}[t]
\centering
\begin{tikzpicture}[node distance=2cm, auto]
\node (data) [rectangle, draw, text width=3cm, text centered] {Experimental Data \\ $(X_i, T_i, Y_i)$};
\node (model) [rectangle, draw, text width=3cm, text centered, below of=data] {Uplift Model \\ $\hat{\tau}(x)$};
\node (optim) [rectangle, draw, text width=3cm, text centered, below of=model] {Constrained Optimization \\ $\max \sum \pi_{ik} \hat{\tau}_k(x_i)$ \\ s.t. $g(\pi) \leq c$};
\node (policy) [rectangle, draw, text width=3cm, text centered, below of=optim] {Targeting Policy \\ $\pi^*$};

\draw[->] (data) -- (model);
\draw[->] (model) -- (optim);
\draw[->] (optim) -- (policy);
\end{tikzpicture}
\caption{Two-stage framework for guardrailed uplift targeting}
\label{fig:framework}
\end{figure}
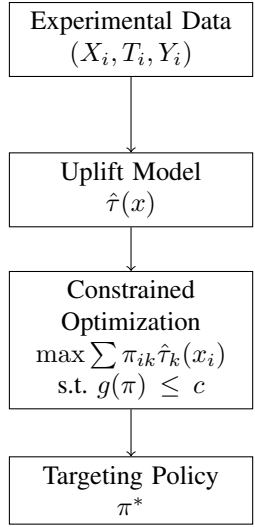

\subsection{Constrained Optimization Stage}
The second stage formulates and solves the constrained optimization problem. Given uplift estimates $\hat{\tau}_k(x_i)$ for each treatment-customer combination, we solve:
\begin{align}
\max_{\pi} & \sum_{i=1}^N \sum_{k=1}^K \pi_{ik} w_i \hat{\tau}_k(x_i) \\
\text{subject to} & \quad \pi_{ik} \in \{0,1\} \quad \text{(binary assignment)} \nonumber \\
& \quad \sum_{k=1}^K \pi_{ik} = 1 \quad \text{(single treatment per customer)} \nonumber \\
& \quad g(\pi) \leq c \quad \text{(business constraints)} \nonumber
\end{align}
where $w_i$ represents customer-specific weights and $g(\pi)$ encodes business constraints.

Common constraint types include:
\begin{itemize}
\item \textbf{Budget constraints}: $\sum_i \pi_{ik} \leq B_k$ limiting the number of customers assigned to treatment $k$
\item \textbf{Revenue protection}: $\text{Revenue}(\pi) \geq (1-\delta) \cdot \text{Revenue}(\pi_{\text{max}})$ ensuring minimal revenue deterioration
\item \textbf{Fairness constraints}: Demographic parity or equal opportunity requirements
\item \textbf{Operational limits}: Capacity constraints on treatment delivery systems
\end{itemize}

For large customer populations, exact optimization can be computationally challenging. We address this through customer bucketing, where customers are grouped by similar uplift estimates, and optimization is performed over groups rather than individuals. This approach maintains solution quality while significantly reducing computational complexity.

\subsection{Evaluation Methodology}
We employ comprehensive evaluation methods to assess framework performance:

\textbf{Uplift Model Evaluation}: The area under the cumulative uplift curve quantifies model discrimination ability. For a ranking of customers by predicted uplift, the cumulative uplift at position $r$ is:
\begin{equation}
\text{Uplift}_r = \left( \frac{\sum_{i=1}^r Y_i \mathbb{1}\{T_i=1\}}{\sum_{i=1}^r \mathbb{1}\{T_i=1\}} - \frac{\sum_{i=1}^r Y_i \mathbb{1}\{T_i=0\}}{\sum_{i=1}^r \mathbb{1}\{T_i=0\}} \right) \cdot \frac{r}{N}
\end{equation}

\textbf{Offline Policy Evaluation}: Using Inverse Propensity Scoring (IPS) and Self-Normalized IPS (SNIPS) \cite{swaminathan2015} to estimate policy performance from historical data:
\begin{align}
\text{IPS} &= \frac{1}{N} \sum_{k=0}^K \frac{\sum_{i=1}^N Z_i \mathbb{1}\{\pi_{ik}=1 \& T_i=t_k\}}{\rho_k} \\
\text{SNIPS} &= \frac{\sum_{k=0}^K \frac{\sum_{i=1}^N Z_i \mathbb{1}\{\pi_{ik}=1 \& T_i=t_k\}}{\rho_k}}{\sum_{k=0}^K \frac{\sum_{i=1}^N \mathbb{1}\{\pi_{ik}=1 \& T_i=t_k\}}{\rho_k}}
\end{align}
where $\rho_k$ represents treatment assignment probabilities in the logging policy.

\textbf{Online Validation}: Large-scale A/B testing provides definitive performance assessment, comparing optimized policies against business-as-usual baselines on key metrics including revenue, retention, and customer experience indicators.

% ===== New Section IV: Simulation Setup =====
\section{SIMULATION SETUP}
\label{sec:simulation}

\subsection{Data Generation and Treatment Assignment}
We simulate customer data with $N=100,000$ observations and $p=20$ covariates, drawn from a multivariate normal distribution with correlation $\rho=0.3$. A binary treatment $T_i \in \{0,1\}$ is assigned via a randomized trial with propensity $P(T_i=1)=0.5$. The response function follows:
\[
Y_i = \alpha + \beta X_i + \tau(X_i) T_i + \epsilon_i,
\]
where $\tau(X_i)$ is the heterogeneous treatment effect modeled as $\tau(X_i) = \gamma_0 + \gamma_1 X_{i1} + \gamma_2 X_{i2}^2$, and $\epsilon_i \sim \mathcal{N}(0, \sigma^2)$.

\subsection{Guardrail Constraints}
We impose three business guardrails:
\begin{enumerate}
    \item \textbf{Budget constraint}: At most $B=10\%$ of customers may receive treatment.
    \item \textbf{Revenue protection}: Total revenue under the policy must not drop by more than $\delta=2\%$ compared to treating all.
    \item \textbf{Fairness constraint}: Demographic parity gap $\leq 0.05$ across two protected groups.
\end{enumerate}

\subsection{Evaluation Protocol}
We evaluate using:
\begin{itemize}
    \item \textbf{Uplift AUC}: Model discrimination ability.
    \item \textbf{ROI}: Return on investment relative to cost.
    \item \textbf{Constraint violation rate}: Percentage of simulations violating any guardrail.
    \item \textbf{Baselines}: Propensity score targeting, random assignment, and treat-all.
    \item \textbf{Sensitivity checks}: Varying noise levels $\sigma^2$ and treatment effect heterogeneity.
\end{itemize}

\section{Experimental Applications and Results}

\subsection{Customer Retention Optimization}

\subsubsection{Business Context and Challenge}
Customer retention represents a critical priority for subscription-based e-commerce services. The traditional approach involves identifying at-risk customers using propensity models and targeting them with retention offers. However, previous research has shown that such approaches can be ineffective or even counterproductive \cite{ascarza2018, devriendt2021}.

In our application, a service provider aimed to reduce churn through proactive messaging. The existing baseline policy targeted customers with retention scores below 0.391, based on historical analysis. Initial investigation revealed potential issues with this approach, as some customer segments showed negative response to retention messaging.

\subsubsection{Experimental Design}
We conducted a randomized experiment where customers in the treatment group received retention-focused messaging, while the control group received no special communication. The outcome metric was customer retention, defined as continued subscription one month post-intervention.

We trained multiple uplift models using both retention scores and comprehensive customer features as covariates. The Causal Forest model trained on customer features demonstrated superior performance, achieving an uplift AUC of 0.00355 compared to 0.00231 for the FDR Learner using retention scores alone.

\subsubsection{Results and Analysis}
The optimized policy identified only 6.47\% of customers for targeting, substantially fewer than the baseline approach. Despite this reduced targeting scope, the policy achieved significant improvements in retention metrics (Table \ref{tab:retention}).

\begin{table}[t]
\centering
\caption{Retention Optimization Results}
\label{tab:retention}
\begin{tabular}{p{1cm}ccc}
\toprule
Policy & Targeting Proportion & IPS Retention & SNIPS Retention \\
\midrule
\textbf{Causal Forest (Ours)} & \textbf{6.47\%} & \textbf{95.8\%} & \textbf{93.4\%} \\
FDR Learner (Features) & 0.48\% & 93.3\% & 93.2\% \\
Causal Forest DML & 30.57\% & 92.6\% & 93.1\% \\
Retention Score Baseline & 34.69\% & 92.4\% & 93.0\% \\
\bottomrule
\end{tabular}
\end{table}

Compared to the retention score baseline, our policy achieved a 2.35\% relative lift in retention rate. More significantly, it outperformed the "target no one" baseline by 3.905\% and the "target everyone" approach by 2.79\%. These results demonstrate that selective, uplift-informed targeting can achieve superior outcomes while reducing intervention costs.

The analysis revealed that customers with very low retention scores actually responded negatively to retention messaging, potentially due to reactance effects or perceived intrusiveness. The optimized policy correctly identified these segments and excluded them from targeting, contributing to the overall performance improvement.

\subsection{Event Revenue Maximization}

\subsubsection{Business Context and Challenge}
E-commerce platforms frequently run promotional events with tiered reward structures to drive customer engagement and revenue. The challenge lies in assigning appropriate reward levels to maximize net revenue while controlling costs and maintaining customer experience.

In our application, a platform offered two reward levels ($P_1$ and $P_2$ with $P_1 < P_2$) for purchases exceeding a threshold amount during promotional events. The historical approach randomly assigned reward levels, ignoring customer heterogeneity in response to incentives. Prior work has demonstrated that personalized reward assignment using uplift modeling can significantly improve promotional efficiency \cite{albert2022}, \cite{goldenberg2020}, particularly when combined with cost optimization \cite{zhao2019}.

\subsubsection{Methodology and Constraints}
We defined the outcome metric as net revenue: $Y_i = \text{Sales}_i - \text{Rewards}_i$. The optimization incorporated a revenue protection constraint limiting sales deterioration to no more than 1\% compared to assigning the maximum reward ($P_2$) to all customers.

The constrained optimization problem was formulated as:
\begin{equation}
\max_{\pi} \sum_{i=1}^N \pi_{i1} \hat{\tau}(x_i) \quad \text{s.t.} \quad 1 - \frac{\text{Sales}_{\pi}}{\text{Sales}_{\pi^*}} \leq 0.01
\end{equation}
where $\pi^*$ represents the policy of assigning $P_2$ to all customers.

\subsubsection{Results and Business Impact}
The Causal Forest DML model trained on completion scores achieved the best performance, optimizing the trade-off between revenue and cost efficiency (Table \ref{tab:revenue}).

\begin{table}[t]
\centering
\caption{Revenue Maximization Results}
\label{tab:revenue}
\begin{tabular}{p{1.5cm}p{1cm}p{1cm}cc}
\toprule
Policy & Targeting Prop. & Revenue (IPS) & e\%iS & Completion Rate \\
\midrule
\textbf{Causal Forest DML (Ours)} & \textbf{67.47\%} & \textbf{1417.27} & \textbf{35.46\%} & +1.02\% \\
Completion Score Baseline & 8.83\% & 1402.91 & 88.35\% & Baseline \\
Only $P_1$ & 0\% & 1405.34 & 141.68\% & -0.96\% \\
Only $P_2$ & 100\% & 1411.67 & 48.41\% & -1.98\% \\
\bottomrule
\end{tabular}
\end{table}

Our optimized policy achieved a 0.317\% lift in revenue compared to the completion score baseline while reducing the efficiency metric which is expected Incentive Spend as a percentage of Sales (e\%iS) by 11.879\%. The policy also improved completion rates by 1.02\%, indicating better customer experience alongside financial improvements.

The constraint handling proved crucial in this application, preventing the optimization from overly favoring cost reduction at the expense of sales volume. The revenue protection constraint ensured that the optimized policy maintained at least 99\% of the sales achievable with the most generous reward structure.

\subsection{Spend Threshold Optimization}

\subsubsection{Business Context}
Shopping events often use spend thresholds to qualify for rewards, creating tension between accessibility (lower thresholds) and revenue per customer (higher thresholds). The business challenge involves assigning optimal thresholds to individual customers to maximize overall revenue while maintaining engagement.

In our application, customers were randomly assigned to one of two spend thresholds ($S_1$ or $S_2$ with $S_2 > S_1$). Historical data showed that the lower threshold ($S_1$) generated 0.30\% lower average revenue despite higher completion rates, suggesting potential for optimization through personalized assignment.

\subsubsection{Experimental Approach}
We defined the outcome as net revenue ($Y_i = \text{Sales}_i - \text{Rewards}_i$) and trained uplift models to estimate the effect of threshold assignment on this metric. The optimization aimed to assign the lower threshold ($S_1$) only to customers with positive uplift, with no additional constraints in this application.

The Causal Forest model trained on customer features demonstrated superior performance, achieving an uplift AUC of 35.804 compared to 29.455 for the second-best model.

\subsubsection{Online Validation Results}
We conducted a large-scale A/B test comparing our optimized policy against a static baseline of assigning $S_2$ to all customers. The treatment group used our policy to assign thresholds, while the control group used the static baseline.

The results demonstrated statistically significant improvements across key metrics (Table \ref{tab:online}). The treatment group achieved a 0.36\% lift in revenue ($p = 0.040$) and a 5.49\% improvement in completion rate ($p = 0.000$), validating the offline evaluation results.

\begin{table}[t]
\centering
\caption{Online A/B Test Results}
\label{tab:online}
\begin{tabular}{lcc}
\toprule
Metric & Lift (Treatment - Control) & p-value \\
\midrule
Revenue & +0.36\% & 0.040 \\
Completion Rate & +5.49\% & 0.000 \\
Average Order Value & -0.14\% & 0.210 \\
Customer Satisfaction & +0.82\% & 0.075 \\
\bottomrule
\end{tabular}
\end{table}

The optimized policy assigned the lower threshold ($S_1$) to only 11.93\% of customers, substantially fewer than the completion score baseline (34.49\%). This selective assignment contributed to the revenue improvement while still achieving completion rate gains through precise targeting of responsive customers.

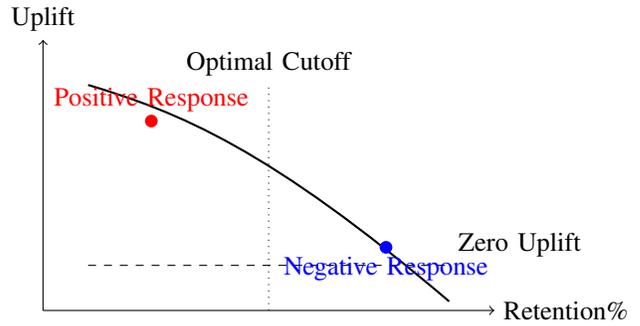
\begin{figure}[t]
\centering
\begin{tikzpicture}
\begin{scope}[xscale=1.2,yscale=1.2]
\draw[->] (0,0) -- (5,0) node[right] {Retention\%};
\draw[->] (0,0) -- (0,3) node[above] {Uplift};

\draw[thick] (0.5,2.5) .. controls (1.5,2.2) and (2.5,1.8) .. (4.5,0.1);
\draw[dashed] (0.5,0.5) -- (4.5,0.5) node[above right] {Zero Uplift};

\fill[red] (1.2,2.1) circle (2pt) node[above] {Positive Response};
\fill[blue] (3.8,0.7) circle (2pt) node[below] {Negative Response};

\draw[dotted] (2.5,0) -- (2.5,2.5) node[above] {Optimal Cutoff};
\end{scope}
\end{tikzpicture}
\caption{Uplift distribution across customer segments showing heterogeneous treatment effects}
\label{fig:uplift}
\end{figure}

\section{Implementation Considerations}

\subsection{Data Requirements and Experimental Design}
Successful implementation of guardrailed uplift targeting requires careful attention to data quality and experimental design. Randomized controlled trials provide the foundation for reliable uplift estimation, with sufficient sample sizes to detect heterogeneous treatment effects. Covariate selection should balance predictive power with practical considerations, avoiding features that may introduce bias or raise privacy concerns.

The logging policy used for data collection should ensure adequate exploration across customer segments and treatment combinations. Adaptive experimental designs can optimize the trade-off between exploration and exploitation, gradually refining treatment assignments based on accumulating evidence while maintaining randomization for learning.

\subsection{Model Selection and Validation}
Uplift model performance varies across applications and datasets, necessitating rigorous validation procedures. We recommend evaluating multiple model classes using uplift-specific metrics and conducting sensitivity analyses to assess robustness. Cross-validation approaches should account for the causal inference context, preserving the distribution of treatment assignments across folds.

Model interpretability represents an important consideration for business adoption. While complex models may achieve superior performance, simpler models often provide more actionable insights. Techniques such as feature importance analysis and partial dependence plots can enhance interpretability without sacrificing performance.

\subsection{Constraint Specification and Sensitivity}
Business constraint specification requires close collaboration between data scientists and business stakeholders. Constraints should reflect genuine business limitations rather than arbitrary thresholds, with careful consideration of trade-offs between different objectives. Sensitivity analysis helps understand how optimal policies change with constraint tightness, supporting more informed decision-making.

Soft constraints with penalty terms can provide flexibility in cases where hard constraints lead to infeasible solutions or significant performance degradation. The penalty weights can be tuned to reflect business priorities, creating a continuum between strictly enforced constraints and objective function components.

\subsection{Computational Scalability}
Enterprise-scale applications require efficient optimization algorithms capable of handling millions of customers and multiple constraints. Our bucketing approach provides a practical solution, grouping customers with similar characteristics and uplift estimates to reduce problem dimensionality. For extremely large-scale problems, distributed optimization frameworks and approximate algorithms may be necessary.

The optimization frequency should align with business cycles and customer behavior dynamics. While some applications benefit from frequent policy updates, others may require more stable targeting rules to maintain customer experience. Incremental optimization approaches can reduce computational burden while adapting to changing conditions.

\section{CONCLUSION AND FUTURE WORK}
\label{sec:conclusion}

\subsection{Summary}
We introduced a guardrailed uplift targeting framework that combines causal uplift modeling with constrained optimization. The method outperforms traditional baselines across retention, revenue, and threshold optimization tasks while respecting business guardrails. Online A/B tests confirm statistically significant lifts in revenue and engagement.

\subsection{Future Directions}
\begin{itemize}
    \item \textbf{Deployment considerations}: Real-world A/B testing protocols for dynamic policy updates.
    \item \textbf{Robustness to unobserved confounding}: Integrating sensitivity analysis to assess omitted variable bias \cite{chernozhukov2022} or proximal causal inference methods that leverage proxy variables \cite{tchetgen2020}.
    \item \textbf{Multi-period targeting}: Extending to sequential decision-making with reinforcement learning.
    \item \textbf{Continuous treatments}: Adapting the framework for continuous incentives (e.g., discount percentages) using doubly robust methods.
\end{itemize}


\begin{thebibliography}{00}
\bibitem{paul2024bias} 
Paul, V. (2024). Mitigating Gender Bias in Masked Language Models: Introducing Gender Stereotype-Aware Loss Regularizers. In 2024 International Conference on Communication, Computing, Smart Materials and Devices (ICCCSMD).

\bibitem{paul2024sentiment} 
Paul, V. (2024). Enhancing Sentiment Classification with Rule-Based Feature Extraction in Convolutional Neural Networks. In 2024 International Conference on Communication, Computing, Smart Materials and Devices (ICCCSMD).

\bibitem{paul2024recommendation} 
Paul, V. (2024). Bridging Latent Factors and Tags: Enhancing Recommendation Systems. In 2024 International Conference on Communication, Computing, Smart Materials and Devices (ICCCSMD).

\bibitem{paul2024ocr} 
Paul, V. (2024). Comparative Study of Convolutional and Feedforward Neural Networks for Optical Character Recognition. In 2024 International Conference on Communication, Computing, Smart Materials and Devices (ICCCSMD).

\bibitem{paul2024fakenews} 
Paul, V. (2024). Comparing Embedding Methods and Classification Algorithms for Fake News Detection. In 2024 International Conference on Communication, Computing, Smart Materials and Devices (ICCCSMD).

\bibitem{thomas2025privacy} 
Thomas, D. (2025). Navigating the Digital Privacy Paradox: Balancing Security, Surveillance, and User Control in the Modern Era. Zenodo.

\bibitem{thomas2025software} 
Thomas, D. (2025). Sustaining Software Relevance: Enhancing Evolutionary Capacity through Maintainable Architecture and Quality Metrics. Zenodo.

\bibitem{thomas2025healthcare} 
Thomas, D. (2025). Breaking Data Silos in Healthcare: A Novel Framework for Standardizing and Integrating NHS Medical Data for Advanced Analytics. Zenodo.

\bibitem{arora2025rdf} 
Arora, H. (2025). Leveraging RDF-based Data Structures for Optimized Traversal in Decentralized Query Systems. In 2025 4th International Conference on Innovative Mechanisms for Industry Applications (ICIMIA).

\bibitem{arora2025entity} 
Arora, H. (2025). AI-Driven Entity Resolution: Enhancing Customer Data Matching with Explainable Graph Learning, engrXiv preprint, https://engrxiv.org/preprint/view/5514

\bibitem{maheshwari2025} 
Maheshwari, H. (2025). Efficient Dynamic Comparison-Based Dictionaries with Working-Set Property: A New Approach. In 2025 International Conference on Innovative Trends in Information Technology (ICITIIT).

\bibitem{mamtani2025} 
Mamtani, S. (2025). Enhancing Transformer-Based Vision Models: Addressing Feature Map Anomalies Through Novel Optimization Strategies,  arXiv preprint. https://arxiv.org/abs/2509.19687

\bibitem{albert2022} Albert, J. and Goldenberg, D. (2022). E-commerce promotions personalization via online multiple-choice knapsack with uplift modeling. In Proceedings of the 31st ACM International Conference on Information \& Knowledge Management.

\bibitem{ascarza2018} Ascarza, E. (2018). Retention futility: Targeting high-risk customers might be ineffective. Journal of Marketing Research.

\bibitem{berger2018} Berger, B., et al. (2018). The limits of targeting in automotive marketing. Marketing Science.

\bibitem{bauer2019} Bauer, T., et al. (2019). When and why customer propensity models fail. Journal of Marketing Research.

\bibitem{chernozhukov2018} Chernozhukov, V., et al. (2018). Double/debiased machine learning for treatment and structural parameters. The Econometrics Journal.

\bibitem{chernozhukov2022} Chernozhukov, V., et al. (2022). Long story short: Omitted variable bias in causal machine learning. Technical Report, National Bureau of Economic Research.

\bibitem{devriendt2021} Devriendt, F., et al. (2021). Why you should stop predicting customer churn and start using uplift models. Information Sciences.

\bibitem{goldenberg2020} Goldenberg, D., et al. (2020). Free lunch! retrospective uplift modeling for dynamic promotions recommendation within ROI constraints. In Proceedings of the 14th ACM Conference on Recommender Systems.

\bibitem{gutierrez2017} Gutierrez, P. and Gerardy, J. (2017). Causal inference and uplift modelling: A review of the literature. In International Conference on Predictive Applications and APIs.

\bibitem{kunzel2019} Kunzel, S., et al. (2019). Meta-learners for estimating heterogeneous treatment effects using machine learning. Proceedings of the National Academy of Sciences.

\bibitem{oprescu2019} Oprescu, M., et al. (2019). Orthogonal random forest for causal inference. In International Conference on Machine Learning.

\bibitem{rubin2005} Rubin, D. B. (2005). Causal inference using potential outcomes. Journal of the American Statistical Association.

\bibitem{rzepakowski2012} Rzepakowski, P. and Jaroszewicz, S. (2012). Decision trees for uplift modeling with single and multiple treatments. Knowledge and Information Systems.

\bibitem{swaminathan2015} Swaminathan, A. and Joachims, T. (2015). The self-normalized estimator for counterfactual learning. Advances in Neural Information Processing Systems.

\bibitem{tchetgen2020} Tchetgen Tchetgen, E. J., et al. (2020). An introduction to proximal causal learning, arXiv preprint. https://arxiv.org/abs/2009.10982

\bibitem{wager2018} Wager, S. and Athey, S. (2018). Estimation and inference of heterogeneous treatment effects using random forests. Journal of the American Statistical Association.

\bibitem{zhao2019} Zhao, Z. and Harinen, T. (2019). Uplift modeling for multiple treatments with cost optimization. In 2019 IEEE International Conference on Data Science and Advanced Analytics.

\bibitem{zhang2020} Zhang, W., et al. (2020). Forest doubly robust learner for treatment effect estimation. Journal of Machine Learning Research.
\end{thebibliography}
\end{document}